\documentclass{article}

\usepackage{graphicx}
\graphicspath{./Figures/}
\usepackage{amsmath} 
\usepackage{siunitx}
\usepackage{parskip}
\usepackage{float}
\usepackage{lineno}
\usepackage{setspace}
\usepackage{ragged2e}
\usepackage{microtype}

\usepackage[english]{babel}

\usepackage[letterpaper,top=2cm,bottom=2cm,left=3cm,right=3cm,marginparwidth=1.75cm]{geometry}

\begin{document}

\onehalfspacing
\justifying
\setlength{\parindent}{0pt}

\title{\color{black} Receptogenesis in a Vascularized Robotic Embodiment}

\maketitle

\author{Kadri-Ann Pankratov*,}
\author{Leonid Zinatullin,}
\author{Hans Priks,}
\author{Adele Metsniit,}
\author{Urmas Johanson,}
\author{Tarmo Tamm,}
\author{Alvo Aabloo,}
\author{Edoardo Sinibaldi*,}
\author{Indrek Must*}

K. A. P., L. Z., H. P., A. M., U. J., Prof. T. T., Prof. A. A., Assoc.Prof. I. M.\\
Institute of Technology, University of Tartu, Nooruse 1, 50411 Tartu, Estonia\\

E. S.\\
Italian Institute of Technology, Via Morego, 30, 16163 Genova, Italy\\

Email Addresses: kadri-ann.pankratov@ut.ee, edoardo.sinibaldi@iit.it, indrek.must@ut.ee


\newpage

\begin{abstract}

\color{black}

Equipping robotic systems with the capacity to generate \textit{ex novo }hardware during operation extends physical adaptability. Unlike modular systems that rely on discrete component integration pre- or post-deployment, we envision physical adaptation through continuous in-body development via hardware synthesis. Drawing inspiration from circulatory systems that redistribute mass and function in biological organisms, we utilize fluidics to restructure the material interface, a capability currently unmatched in robotics. Here, we realize this proof-of-concept hardware generation through a vascularized robotic composite designed for programmable material synthesis, demonstrated via receptogenesis - the on-demand construction of sensors. By coordinating the fluidic transport of precursors with external localized UV irradiation, we drove an \textit{in situ} photopolymerization that chemically reconstructed the vasculature from the inside out. This reaction converted precursors with photolatent initiator into a solid dispersion of UV-sensitive polypyrrole in PETG, establishing a sensing modality validated by a characteristic decrease in electrical impedance. The newly synthesized sensor closed a local control loop in real time to regulate wing flapping in a moth-inspired robotic demonstrator. Our work is a proof-of-concept materials basis for \textit{ex novo} hardware generation in a vascularized composite - a step towards situated robots adapting to environmental cues.

\end{abstract}

\section{Introduction}

\color{black} 

In unstructured environments, critical physical signals are not static features that can be precisely registered by a dedicated sensor, but rather emergent phenomena arising from the robot's unique interactions with the environment \cite{Pfeifer2007Self-OrganizationRobotics, Alu2025RoadmapStructures}. We need more focus on situated robots \cite{Vihmar2025Silk-inspiredRobots} that are deeply intertwined with their immediate surroundings and leverage the environment itself to guide their behaviour and expand their physical capabilities. In situ generation of new hardware within the robot body based on environmental cues would couple environmental information with physical updates, adaptively creating structures that transduce situation-specific signals emerging from the environment.

Prior approaches to physical adaptation during deployment differ in the extent of bodily changes. Some rearrange existing hardware — through compliance and deformation \cite{Rus2015DesignRobots}, appendage repurposing for locomotion plasticity \cite{Sihite2023Multi-ModalEnhancement}, shape-morphing limbs \cite{Baines2022Multi-environmentMorphogenesis} or body-wide morphological reconfiguration \cite{Polzin2025RoboticEnvironments} for multi-environment transitions, or software remapping of existing sensors, e.g., repurposing navigational accelerometers for gravimetry in the Curiosity rover \cite{Lewis2019ACrater}. Self-healing systems restore the body after damage, knitting severed material back together through reversible bonds in embedded polymer networks \cite{Terryn2020RoomNetwork, Bai2022AutonomousSystems}; the endpoint, however, is a recovered prior state rather than a new function. Growth-based robots incorporate genuinely new matter, as in root-inspired systems that bind ambient soil with a tip-deposited linker \cite{Sachin2025BeeRootBot:Binding}; here, material is accreted at an external growth front rather than synthesized within the existing body. Retrofitting strategies place new functional components onto the robot via in-situ 3D printing \cite{Kanhere2024UpgradingPrinting}, or repurpose existing fluidic channels to extract sensory information from a single pressure input \cite{Zou2024ARobots}; in both cases, the added capacity remains structurally discrete from the body it extends, acting as an add-on rather than an integral part that invokes new behavior. Across all these approaches, the new capacity is either a rearrangement of what was already there, a return to the prior state, or an external appendage — none synthesize hardware that is structurally and functionally continuous with the body, formed internally, and providing a new qualitative function.

\begin{figure}[H]
    \centering
    \includegraphics[width=\linewidth]{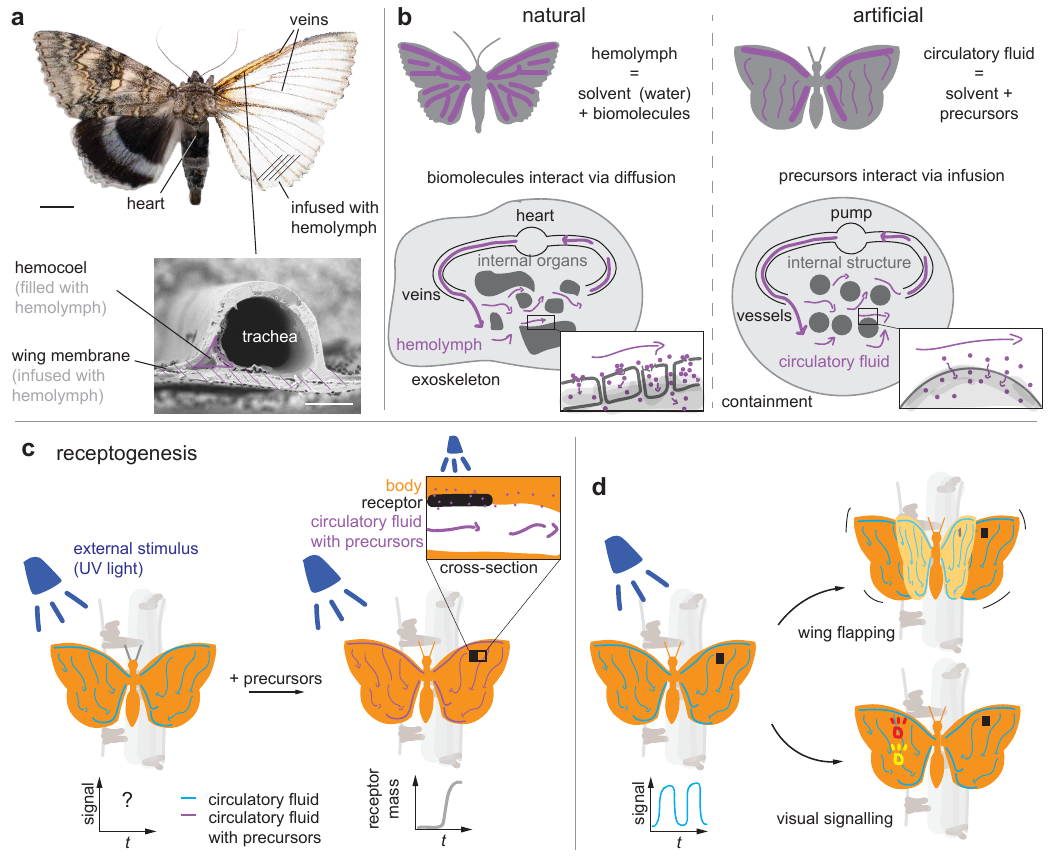}
    \caption{\textit{Ex novo} hardware (receptor) generation in a moth-inspired vascular system: system concept a) Biological model. Dorsal view on \textit{Catocala fraxini}, scales removed on the right wing to visualize wing veins. The SEM image inset shows a cross-section of a wing vein for carrying hemolymph.  b) System components of a natural and artificial open circulatory system, respectively. Connectedness is achieved with directing fluid flow and interaction with internal elements. c) Receptogenesis concept. Precursors introduced on demand to the vasculature form a receptive area in response to an externally applied stimulus. d) The local physical update gives access to receptor data (left) that closes emergent control loops, exemplified by regulating the flapping of artificial wings (top right) or visual signaling (bottom right). Scale bars (a) - \SI{1}{\centi\meter}, (a, inset) - \SI{100}{\micro\meter}}
    \label{fig:1}
\end{figure}

Synthesizing hardware internally to the body shifts the problem from component placement to coupled transport and assembly: precursors must travel centimeters through a vascular network, then convert into function across micrometres through molecular assembly. In nature, analogous material transport issues are solved with passive mechanisms (diffusion) for molecule-level interactions such as receptor binding, and active mechanisms (advection, particularly circulation) to minimize diffusion distance \cite{Monahan-Earley2013EvolutionaryEndothelium}. In robotics, the open circulatory systems of simpler organisms such as arthropods serve as architectural models for material transport networks. They consist of a pumping element (heart), flow-directing arteries, and hemolymph — a circulatory fluid that bathes the internal organs and infuses the wings, as in Lepidoptera (Figure \ref{fig:1}a). Insect wings are living structures \cite{Salcedo2023ComplexWings}, in which hemolymph supports cuticle hydration for stiffness modulation \cite{Pass2018BeyondFunctionality}, thermal regulation \cite{Tsai2020PhysicalButterflies}, aerodynamic properties \cite{Pass2018BeyondFunctionality, Brasovs2023HaemolymphFlight}, and chemical \cite{Ehlers2023TheButterflies} and mechanical \cite{Pratt2017NeuralWings} signalling, providing — with effectively 2D fluid flow — an accessible case study for vascularised robotics.

Artificial interpretations of vascularization are highlighted by progress in microfluidics, e.g., organ-on-a-chip or drug testing solutions \cite{Ayuso2022AMedicine},  as well as structures for liquid transport \cite{Dudukovic2021CellularFluidics} and patterning \cite{Liu2024BioinspiredTransistors}. In robotic systems, a multifunctional internal liquid was circulated throughout a lionfish-inspired robot, simultaneously powering and actuating it \cite{Aubin2019ElectrolyticRobots}. In terms of intra-fluid communication, Garrad \textit{et al} demonstrated that injecting patterned sequences of conducting and insulating fluid into a soft tube produces resistance changes across embedded electrodes, encoding the fluidic input as a binary electrical signal \cite{Garrad2019ARobots}. In our previous work \cite{Valdur2024ATransport}, we demonstrated the integrative role of internal liquid in simultaneously modulating body compliance and activating muscles, reaching protovascular levels, highlighting its enabling role in physically intelligent robotic systems. For living mammalian cells, oxygen diffusion distance is in the range of \SI{100}{\micro\meter} to \SI{200}{\micro\meter} \cite{Rouwkema2008VascularizationEngineering}, which gives clear motivation to develop vascular structures for cell viability in biohybrid systems \cite{Filippi2022MicrofluidicBio-Actuation} and hints to potential architectures for material-level \textit{in-situ} morphogenesis. Robotic vasculature (also termed embedded functional vias \cite{Baines2024RobotsDemand}) is proposed to alter robots' body properties in response to environmental conditions, offering a pathway to mimic phenotypic plasticity in organisms \cite{Baines2024RobotsDemand}. However, no robotic vasculature has yet realized this plasticity in a local closed loop: from an environmental stimulus, through material change, to a new functional capability of the body.

Taken together, the existing strategies for robots' physical adaptability and vascular architectures have remained disconnected. Vascularization enables through-embodiment augmentation: new functional material is synthesized within the existing body structure, resulting in \textit{ex novo} hardware capability that the robot did not possess at manufacture, without altering the perimeter or compromising structural integrity. When the \textit{ex novo} component is a sensory receptor, it grants the robot access to environmental signals it could not previously detect; we term this \textit{receptogenesis}. Both terms (\textit{ex novo} hardware generation and receptogenesis) describe robots' operational capability, allowing the vast chemical synthesis toolbox to functionally augment robots.

Here, we report the \textit{ex novo} hardware generation through an integrated vascular–materials strategy, realized as receptogenesis. We introduce a two-scale vascularization strategy in which a single fabrication method bridges global and local transport: 3D-printed veins carry precursors across the decimeter-scale body, while infusion drives them into the surrounding PETG matrix down to ~\SI{10}{\micro\meter}. At the component level, we show that polypyrrole (PPy) synthesized via UV-induced photopolymerization is a feasible UV-responsive receptor material. We validate our framework on a moth‑inspired (Figure \ref{fig:1}b, left) flapping‑wing robotic platform (Figure \ref{fig:1}b, right), where a circulatory fluid is routed through an engineered vascular network and used to deliver PPy precursors to the receptor site to be photopolymerized. The newly synthesized receptor closes a control loop regulating wing flapping and visual signaling (Figure 1d), demonstrating a functional capability the robot did not possess prior to deployment. In this paper, we show a single chemical pathway with a predetermined outcome; however, the vascular architecture provides inherent support for a broad range of potential precursors, either dissolved or dispersed in a liquid carrier. This work advances  robotic function in complex environments by extending morphological computation to the molecular scale, and provides a materials‑centric and intrinsic dynamics driven perspective on neurovascular integration in robotic systems.

\newpage

\newpage

\section{Results and Discussions}

\subsection{System Design}

Figure \ref{fig:1}c outlines the key steps of \textit{ex novo} component generation using receptogenesis as a case study. We create functional components \textit{in situ} by circulating precursors through an internal artificial vasculature and photopolymerising them with external stimulus (UV light). Inspired by the open circulatory systems of Lepidoptera (Figure \ref{fig:1}b), we implement a 3D printed artificial vascular network that distributes material precursors throughout the body via global advection and local fluid infusion. We use polypyrrole (PPy) precursors for UV-sensitive receptive area photopolymerization. UV signal increases the conductivity of the newly formed PPy temporarily, which functions as a light‑responsive receptor capable of triggering downstream behaviours such as LED indication or wing flapping (Figure \ref{fig:1}d). In our system, the stimulus that initiates receptogenesis and the signal being detected are incidentally the same, though the approach generalizes to a much broader class of stimuli and signals. Stimulus-induced conductivity changes are known in other systems — e.g., non-percolated conductive-particle composites that fuse into a conductive network under current-driven heating \cite{Allen2008} — but whereas those reconfigure a particle population already present at manufacture, our approach synthesizes the functional material in-volume from precursors delivered on demand through the vasculature. Even when narrowly considering PPy, the case study material in this paper, the synthesis result is based on the precursor choice and environmental (synthesis) conditions, leading to a wide range of possible outcomes (e.g., towards bioelectronic interfacing \cite{Dong2024ElectrochemicallyInterfaces}, ionic sensitivity \cite{Wang2016IonDetection}, corrosion resistance \cite{Cao2025HighlyCoatings}, or electromagnetic wave shielding \cite{Kaynak1994AFilms}). Even considering the light-patterning technique, external stimulus can modulate photocrosslinking patterns for microphase separation of a copolymer, achieving stiffness patterning in a chemically identical material \cite{Qi2025MultimaterialsCopolymer}, providing possible application scenarios in our vascularized system.


This work focuses on the 'small' loop — local, stimulus-mediated synthesis in a vascular system primed to encounter the stimulus. The 'large' loop (central decision-making/autonomy, e.g., priming the body with task-relevant precursors, downstream behavior) is implicit: in the experiments reported here, priming was performed via user input, and the body was then exposed to controlled stimuli that activated the primed receptor site. Within the small loop, no central involvement is required during execution — body articulation, self-occlusions, surface orientation, and ambient light gradients would collectively determine where and when polymerization occurs in a deployed system, a role played in our demonstration by a physical mask. Synthesis is therefore shaped by the conditions of each encounter, intrinsically coupling the generated hardware to the environment in which it operates.

In our work, precursors are a liquid mixture of pyrrole (Py, acting as both monomer and solvent), a bis(4-methylphenyl)iodonium hexafluorophosphate initiator, and cellulose acetate propionate (CAP) as a structural modulator. We conceptualize this chemical inventory as a dual-source resource: it may be integrated as an on-board reservoir to ensure operational readiness, or alternatively, sequestered from the ambient environment via harvesting mechanisms. While the logistics of precursor procurement fall outside the primary scope of this study, the capacity for both intrinsic storage and extrinsic acquisition presents vascularization as a robust framework for system augmentation. We chose polyethylene terephtalate glycol (PETG) as the 3D printing polymer as it is commercially available and transmits UV light starting at \SI{310}{\nano\meter} \cite{Arangdad2019InfluenceFilms}. Pyrrole as an organic solvent infuses PETG and can be photopolymerised into PPy \cite{Rabek1992PolymerizationPhotoinitiators}, during which the transparent Py forms dark UV-sensitive \cite{Luhakhra2023PolaronLight} PPy clusters. The perceived color change evidences the synthesis process.

\begin{figure}[H]
    \centering
    \includegraphics[width=\linewidth]{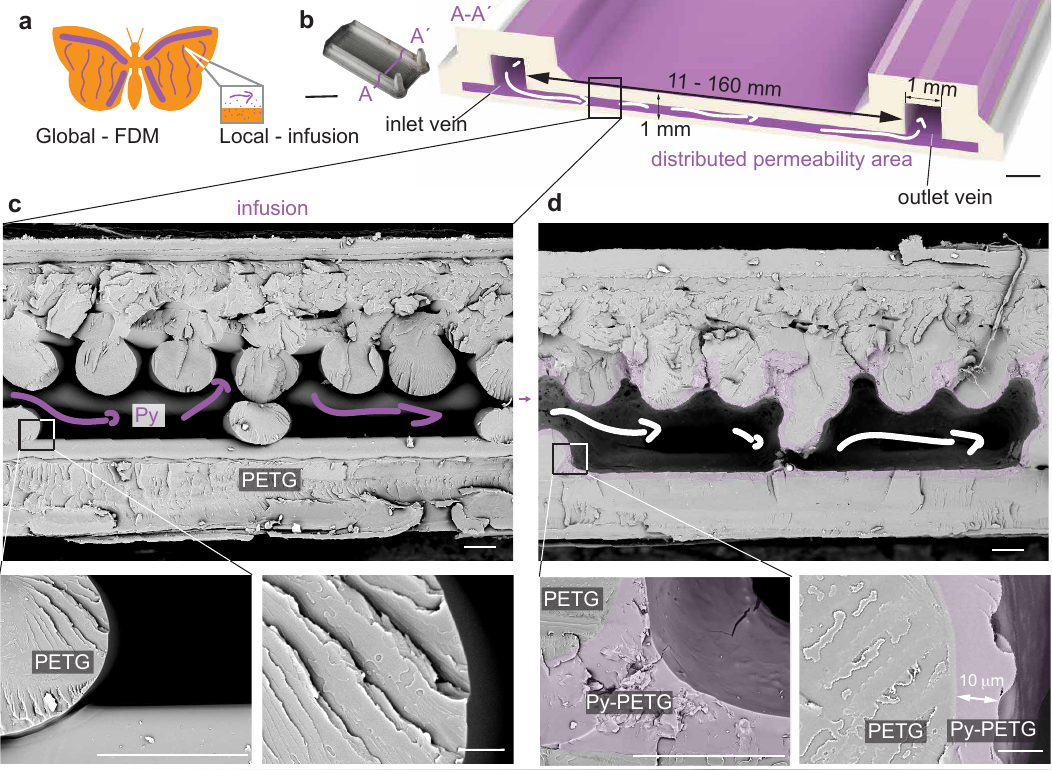}
    \caption{Multiscale vascularization. a) Vascularization hierarchies -  global transport (FDM) and local interaction (infusion). b) Photograph and cross-sectional drawing of a global vascularised robotic composite consisting of veins and microchannels. c) SEM images of a non-infused structure, complementary space between the individual infill lines form a network of microchannels. d) SEM images of an infused (and thereafter dried) structure.  Scale bars: (b, main) - \SI{1}{\milli\meter}, (b, inset) - \SI{10}{\milli\meter}, (c, d, main) - \SI{100}{\micro\meter}, (c, d, insets) - \SI{10}{\micro\meter}}
    \label{fig:2}
\end{figure}

We focus on the physical update (chemical synthesis) in the PETG matrix, opposed to surface-level modifications ("add-ons") that often act as functional obstructions (analogous to intravascular clots), limiting the functional versatility of the vascular system. In macroscale robots, diffusion alone is too slow to deliver precursors uniformly throughout the volume, limiting the full activation rate. To bypass this bottleneck, we implement advection that ensures rapid, uniform activation. We developed a global logistics network combining rapid fluid circulation through a vascular scaffold complemented by slower, diffusion-limited local precursors delivery via infusion into the PETG structure (Figure \ref{fig:2}a). Fused deposition modelling (FDM) printing of PETG produced logistic highways (veins) by explicit definition and a distributed permeability structure (microchannels) by designed infill pattern (Figure \ref{fig:2}b, Supplementary Figure S1, Supplementary Movie 1). In the microchannels (Figure \ref{fig:2}c), the circulatory fluid infused the PETG scaffold (Figure \ref{fig:2}d), enabled by the large surface area in contact with the fluid. Precursor mixture infused the PETG (exposure time approximately \SI{20}{\second}) to an infusion depth of approximately \SI{10}{\micro\meter} as seen from the element contrast in the SEM image in Figure 4d (bottom right). The formed Py-PETG layer was approximately 20\% of the print line's thickness (\SI{50}{\micro\meter} radius). This enables us to achieve connectedness across multiple orders of scale (tens of centimeters via global transport and micrometers via infusion) using a single fabrication method. 

\subsection{Receptogenesis by In-vasculature Molecular Assembly}

\begin{figure}[H]
    \centering
    \includegraphics[width=\linewidth]{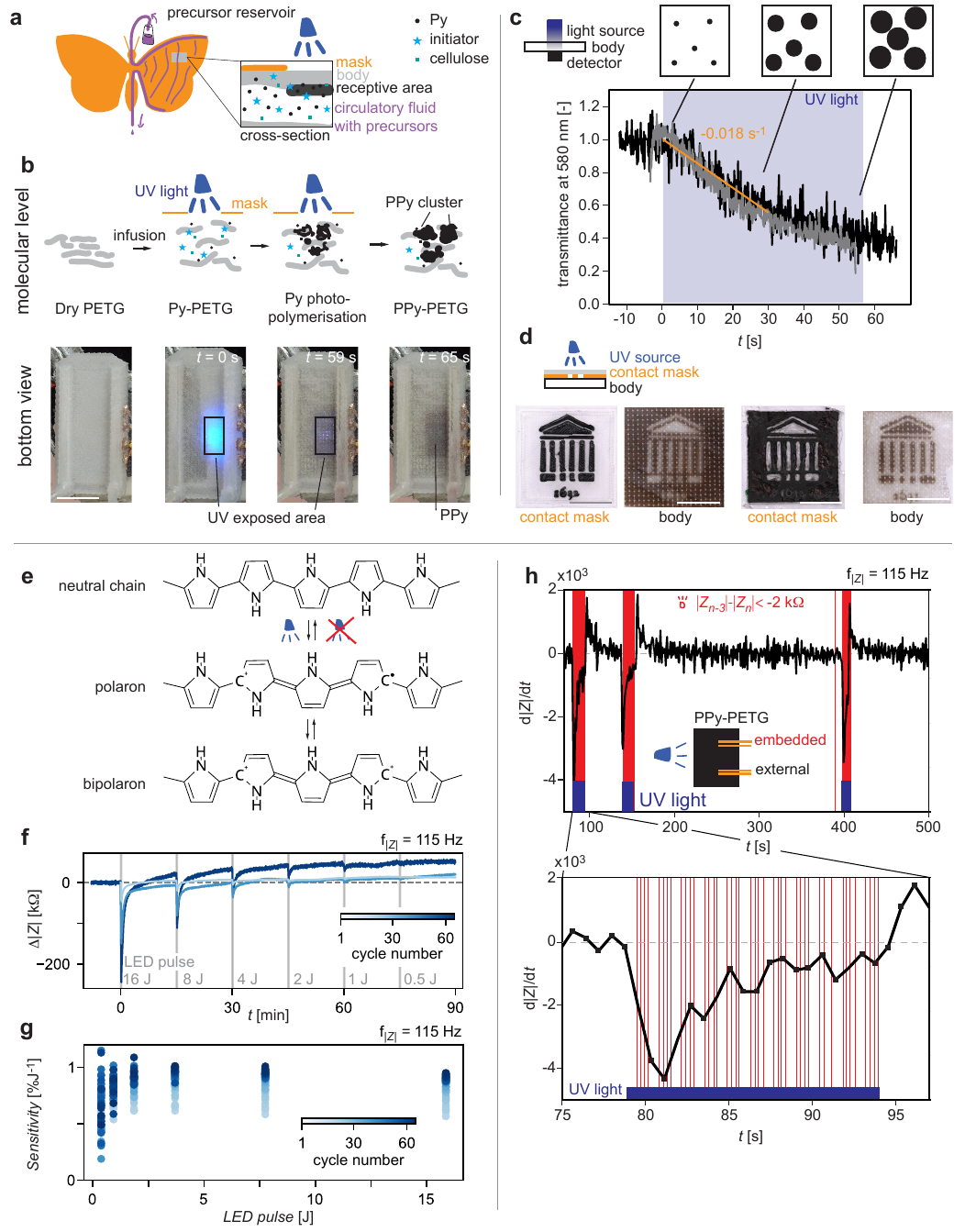}
    \caption{Receptogenesis based on PPy \textit{in situ} photopolymerization a) The receptive area forms by local UV-exposure of the structure with precursors from the chemical inventory and distributed by a vascular system. b) Scheme of PETG infusion with precursors and in-situ photopolymerization stimulated by an external UV light source (at \SI{365}{\nano\meter}) (top). Transilluminative snapshots of photopolymerisation, where PPy synthesis is evidenced by the darkening (bottom). c) \SI{580}{\nano\meter} transmittance through the receptive area decreases during photopolymerization, corresponding to PPy cluster growth (insets). Black and grey lines correspond to two representative photopolymerization instances. d) UV-PPy photolithography demonstration on structure infused 7 days prior. e) UV induced temporary (bi)polaron: reaction scheme. Adapted from \cite{Bredas1985PolaronsPolymers}. f) Cycling stability of the transient differential impedimetric response of the receptive area at varying LED electrical energy input ($\int V \cdot I \, dt \ [\text{J}]$) values of the UV-LED pulses. g) PPy receptive area sensitivity (relative impedance drop divided by LED pulse energy that acts as a proxy for UV dose) increased in cycling. h) Impedimetric transient response (black) of  receptive area to UV irradiation (blue) and modulated optical readout of the receptor (red). Scale bars: (b, d) - \SI{1}{\centi\meter}}
    \label{fig:3}
\end{figure}

Receptogenesis is achieved via photopolymerization of Py precursors into a UV-receptive area induced by initiator decomposition under UV-stimulus (Figure \ref{fig:3}a). The circulatory fluid of precursors infused the PETG structure (Figure \ref{fig:3}b, Supplementary Movie 2) and formed an infused PETG (Py-PETG) region in the structure. Py-PETG formed a solid solution, in which the precursor molecules retained sufficient mobility to reorganize within the matrix, enabling residual polymerization after the initial UV exposure (synthesis event). Since PPy and the precursors absorb UV light, synthesis was localized at UV-exposed surfaces. The reaction area progressively and irreversibly darkened at an absorption rate of \SI{-0.018}{\per\second}, monitored at \SI{580}{\nano\meter} \textit{in-situ} during synthesis (Figure 3c). Based on prior literature of photopolymerization \cite{Rabek1992PolymerizationPhotoinitiators, Kasisomayajula2016ConductivePhotopolymerization} and known PPy morphology from other chemical synthesis methods \cite{Kausaite-Minkstimiene2015EvaluationParticles,Omastova2003SynthesisSurfactants}, PPy most likely formed clusters in the Py-PETG matrix that  self-inhibited further cluster growth when the immediate vicinity has been depleted of precursors. This is consistent with non-zero final transmittance of \SI{580}{\nano\meter} in Figure 3c. Conduction combines electronic conduction of PPy clusters and electrolytic conduction of initiator in the PETG matrix surrounding the clusters, forming unique formations each time, in agreement with concept described elsewhere \cite{Yan2025TailoringEncryption}. Absolute resistance varied widely between samples (\SIrange{200}{2000}{\kilo\ohm}), consistent with sample-specific PPy microstructure, whereas relative UV-sensitivity remained consistent (detailed below). We therefore report UV-responsiveness as a relative resistance change rather than an absolute value. We demonstrated UV-PPy photolithography using a PETG negative contact mask (Figure \ref{fig:3}d) to show that UV-induced synthesis was spatially localized and did not propagate beyond the exposed area, achieving resolution \SI{0.3}{mm}, limited by optical scattering by surface roughness.

As the vicinity of the receptive area retained abundant Py monomers after self-inhibiting PPy photosynthesis, the Py monomers reduced the newly synthesized PPy into an electrically nonconductive form. PPy is sensitive to UV irradiation \cite{Galar2013ChemicalPolypyrrole}: the sensing characteristics of the receptive area (PPy-PETG) derive from the transient formation of (bi)polarons: UV exposure photo-excites PPy by separating charge carriers and modifying the population or dynamics of polaron or bipolaron states (Figure \ref{fig:3}e). The induced positive charge shifts PPy into its oxidized, electrically conductive form, leading to resistance decrease in our PPy-based system. The impedance drop was proportional to the LED pulse (a proxy for UV dose, Figure 3f), indicating that (bi)polaron population was proportional to the incident UV dose. The conductive state is transient, decaying through two routes: charge consumed by further polymerization of the abundant surrounding Py (extending the existing PPy chain), and carrier recombination limited by charge mobility along the chain (timescale $\approx$ \SI{300}{\second}). Over repeated cycling, both sensitivity (Figure 3g) and baseline resistance values increased, consistent with residual polymerization of Py: additional PPy generates more (bi)polarons, increasing sensitivity, whereas a decreased amount of mobile charge carriers increases system resistance. This constant development of the sensor is expected and embodies the desired element of experience to the situated sensor, different from the expectedly constant performance of conventional (industrial) sensors. The system retained functionality after more than four months of storage, demonstrating operational stability on this timescale. For the receptor, the impedance change of the receptive area was registered by an embedded microprocessor unit featuring a custom bipolar pulse technique combined with voltage division by a constant external resistor (Supplementary Figure S2). Further, we encoded receptor response as a differential digital signal with LED readout, given the transient nature of polarons: the receptor response is characterized by its rate of change rather than absolute value, as described above. This parallels biological receptors, as we are not interested in absolute (impedance) values, but rather the change. As seen in Figure \ref{fig:3}h, a drop in impedance corresponds to UV irradiation that was independently visualized with a red LED blinking.

\subsection{Robotic Demonstration}

To demonstrate the efficacy of \textit{ex novo} hardware fabrication for robots in environmental settings, we developed a representative moth-inspired platform (Fig. 4a). This demonstration highlights the synergy between externally applied UV irradiation and body adaptation via material-level response (UV receptivity), which is expressed via behavioral outputs such as visual signalling and wing flapping (mediated by "central nervous system" represented by embedded microprocessors). 

\begin{figure}[H]
    \centering
    \includegraphics[width=\linewidth]{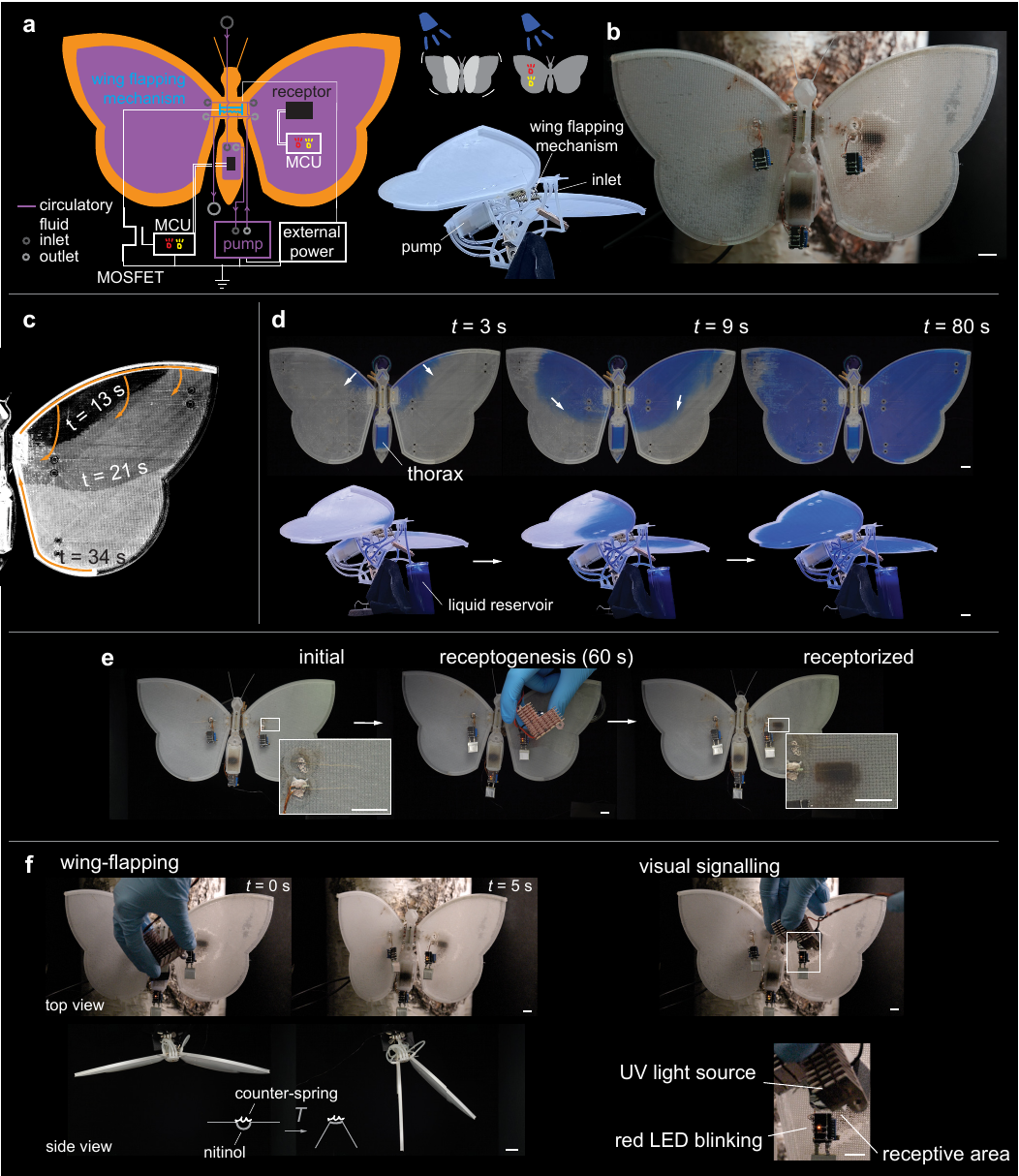}
    \caption{Illustrative robotic demonstration of \textit{ex novo} hardware genesis. a, b) System implementation: a) scheme showing the fluid, control, and power lines and b) photograph. c, d) Illustrative filling of vascular system: c) air displaced with isopropanol  (superimposed timestamps) and d) isopropanol displaced with isopropanol:water with blue tracer (sequential snapshots of top and isometric view) e) Wing-scale receptogenesis. f) The robotic moth reacts to UV-exposure to a receptive area in the thorax section by activation of the wing flapping mechanism and visual indication (LED flashing). Scale bars: \SI{1}{\centi\meter}}
    \label{fig:4}
\end{figure}

UV exposure permanently reduced the material's permeability to photons and unlocked a latent UV-sensing capability, transforming it from a passive substrate into a functional receptive area. The robot started in minimal configuration, with embedded contacts representing latent positions of the receptors in PETG with nonexistent sensitivity to UV stimulus. The embodiment consisted of a thorax (vascular shadow area \SI{3.4}{\square\centi\meter}) and two wings (each with vascular shadow area \SI{86}{\square\centi\meter}, approximately 25 times the size of the thorax, highlighting the scalability of the system) with a total internal vascular volume of approximately \SI{3}{\milli\litre}. The robot's operation represents the co-expression of the central strategic level and the local material-level execution. Upon a central decision (suggesting the emergent need for UV receptivity in the future, not addressed in this work), the robot casts a cocktail of precursors into the vascular system spanning the entire moth-shaped embodiment. In line with the open circulatory systems of animals that expose the entire vasculature to the same circulatory fluid with minimal control and exert its effect locally, here, the whole vasculature was primed for UV encounter, but a receptor developed in irradiated areas only. Indeed, internal circulatory fluid (here, isopropanol, to demonstrate the compatibility of the same vascular system with multiple liquid carriers) reached the entirety of the vascular structure, as seen in the timestamps of dry structure filled with solvent (Figure \ref{fig:4}c, Supplementary Movie 3), and liquid agitation is shown by adding a blue tracer to the solvent  (Figure \ref{fig:4}d, Supplementary Movie 3). The circulatory liquid passed from the leading vein directly into the distributed permeability (vascularized) area,  assisted by pump advection and wetting (capillary forces). Our vascularized system is resilient to many issues, e.g., bubbles, commonly encountered when working with liquids in confined spaces, and uses just a single hydraulic pump with no additional valves or other flow control systems to serve the entire embodiment.

At low UV exposure (\SI{0.9}{\watt} LED at \SI{20}{\milli\meter}), the prepared (Py-PETG, not PPy-PETG) embodiment did not sense UV (Supplementary Figure S3). Upon encounter of the prepared embodiment with intensive UV stimulus (\SI{1.6}{\watt} LED at \SI{43}{\milli\meter}), the embodiment reacted by receptive area synthesis (Figure \ref{fig:4}e, Supplementary Movie 4), which led to UV-responsiveness (sensitive to both \SI{0.9}{\watt} test and \SI{1.6}{\watt} synthesis LED) that completed the control loop by activating wing flapping (electrothermal activation of nitinol by binary MOSFET-controlled heating at \SI{6.6}{\watt} to cause recovery of a preprogrammed closed-wing shape; the flat-wing configuration was recovered by natural convective cooling and spring-assisted return) or visual signalling  (Figure \ref{fig:4}f, Supplementary Movie 5).

\subsection{Discussion and Future Perspective}

Chemistry is an undertapped toolkit in robotics: it opens up a broad range of bodily modifications beyond the combinatory rearrangement of preassembled parts and tightly couples those modifications to the unique conditions under which they occur. This coupling between chemical reaction and environment — bridging molecular dynamics and macroscopic structure — has been proposed as a basis for open-ended computation in natural systems \cite{Gunji2026Quantum-likeEnvironment}, and is already being exploited in multisensory neuromorphic platforms that fuse chemical and visual cues \cite{Zheng2024ACues}. In the same spirit as the present work, Samal \textit{et al.} recently demonstrated in vivo polymerization of n-doped conducting polymers, catalyzed by heme and whole blood, for optically driven neuromodulation in live animals \cite{Samal2026Blood-catalyzedControl} - a striking demonstration that functional electronic hardware can be synthesized inside a living body from circulating precursors. Our work brings this principle into the engineered domain: a vascularized robotic embodiment whose circulating fluid delivers precursors to sites where they are converted, on demand, into the hardware the robot needs (not necessarily centrally aware of the need nor informed of component generation).

The contribution here is one instantiation of a broader strategy: vascular precursor delivery feeding a localized, stimulus-triggered synthesis, in which the chemistry and stimulus are each design choices rather than fixed features. Vascular systems in robotics have previously been demonstrated across a range of distinct functions - energy storage and actuation \cite{Aubin2019ElectrolyticRobots, Wehner2016AnRobots}, communication \cite{Garrad2019ARobots}, thermal regulation \cite{He2019BioinspiredMuscle}, and structural solidification via fluid-phase transitions \cite{Ranzani2018IncreasingSelf-Folding} - each representing a separate instantiation of the same delivery architecture. \textit{Ex novo} hardware generation describes robotic operational capability, not a specific chemistry; prior examples may include in-situ 3D printing \cite{Kanhere2024UpgradingPrinting, Sachin2025BeeRootBot:Binding}. Our work adds to this repertoire by demonstrating the synthesis of a new responsive structure within the body in response to external UV dose rather than relying on internal models, as in 3D printing.

The realization of \textit{ex novo} receptogenesis through vascularized material transport points toward a paradigm where physical adaptation is governed by molecular-scale interactions. In the demonstrated case, the wing structure of the robotic lepidopteran transformed from a translucent, non-conductive state to a black, UV-sensitive "phenotype". We interpret this as offloading the computational burden of bodily modifications onto the interaction between environmental stimuli and molecular structures: conceptually, the synthesis can be viewed as the execution of a program that combines pre-determined bodily affordances (here, the PPy precursors) with situation-specific spatiotemporal information (here, local UV exposure). The newly synthesised hardware is thus a physical record of the environmental conditions. We propose that the augmentation of existing robotic components through site-specific chemical changes, achieved without digital symbolic representation, could help bridge the groundedness gap between digital control and material reality. 

In \textit{ex novo} hardware chemical generation of an internal component, each synthesis event produces a component shaped by the specific conditions of that encounter - environmental light gradients, local chemistry, body geometry - we regard this variation not an experimental shortcoming but a defining feature of the generative architecture. Our design decisions embraced situation-specific, and thus intrinsically non-repeatable, synthesis mechanisms; e.g., we used the relative rate of change rather than the absolute value of impedimetry data as a sensitive area readout in a receptor, precisely because the absolute characteristics of each generated sensor \textit{will} differ and evolve in time. In conventional sensing, such time- and sample-dependent sensitivity would be undesired for analytical applications. Unlike conventional UV sensors that provide instantaneous, memoryless intensity values, this sensing area acts as an intrinsic leaky integrator. After pulse dosing, the material exhibits a slow (\SI{300}{s}) return towards its initial baseline, in part through the extension of receptive PPy chains. The accumulation of new bonds in molecular structure implies that the material's structural topology becomes a tangible record of the agent's physical experience, a behavior we interpret as a form of experiential material memory. This temporal response represents an incremental, state-dependent hardware update characterized by an intrinsic stochasticity component in the final structure. We view it as a step from signal acquisition towards non-von-Neumann neuromorphic material computation. We note that this variability has been exploited as a feature in related work - the same principle of environment-coupled synthesis yielding unique, non-reproducible material signatures has recently been demonstrated as a cryptographic primitive \cite{Yan2025TailoringEncryption}.

The coordination of advection-based global transport and infusion-based local "last-mile delivery" caters to a high surface area and volume. At the subsystem level, the system enables high-density functionalization, suggesting a viable pathway towards organism/life-level intertwined hierarchical functionality that, within the current fabrication-centric paradigm, is challenging to fabricate and nearly impossible to modify or repair post-deployment. The high ratio between the total circulatory surface area and the footprint of individual components allows a centralized fluidic reserve to address and modify a vast number of discrete physical features at once, analogous to a single-wire architecture robustly managing complex electronic arrays. The vascular architecture retains compatibility with microfluidic precision manipulation and targeted delivery approaches; e.g., controlled wetting in the vasculature could target synthesis templated by a central command unit and executed in a distributed manner. Both levels of control - central and local - are needed in a practical system, but currently it is either one or the other, without vertical communication, which handicaps performance. 

As shown in this study, vascularized composites designed to achieve biological levels of adaptability pose challenges, e.g., in tailoring material compositions for suitable robotic preparedness. This challenge is increasingly supported by advancements in computational chemical programming \cite{Leonov2024AnEngine} and machine learning \cite{Duan2025OptimalReactions}, which allow for the design and testing of molecular structures and chemical reactions. The present implementation demonstrates one predesigned chemical pathway, not a true evolutionary or multi-outcome adaptive system; however, by leveraging these computational tools, it becomes possible to design (multiple simultaneous) cascades for stimuli at a specificity similar to that of molecular receptors in biology. Materials capable of robot-wide physical actions establish a robust framework for compositional development and the generation of \textit{ad hoc} hardware in dynamic and unstructured environments.

\section{Conclusion}

In conclusion, we demonstrated the \textit{ex novo} hardware generation (Figure 1) of UV-sensing hardware via photopolymerization within a moth-inspired, vascularized robotic embodiment. By employing a hierarchical vascularization strategy (Figure 2) - combining global advection defined by 3D printed structure with local infusion - we successfully distributed precursors (namely, pyrrole) across the body, enabling effective volume activation upon localised UV-exposure. The resulting receptive area (PPy) exhibited a quantifiable impedance shift due to reversible (bi)polaron formation upon UV exposure (Figure 3). The effective creation of a UV receptor was successfully demonstrated with visual signalling (LED lights) and wing flapping (actuation) (Figure 4). Our work, realized through conventional chemistry and 3D printing, paves the way for a new paradigm for situated robotics, particularly where environmental adaptiveness is required.

\section{Methods}

\textit{Vascularization.}  The vascular structure was designed in Autodesk Fusion, sliced in Orca Slicer, and printed either on Bambu Lab A1 with a \SI{0.2}{\milli \meter} nozzle (printing configuration in Supplementary file 1) or Prusa Research Original Prusa XL Five-Head with a \SI{0.25}{\milli \meter} nozzle with a built-in load cell probe (printing configuration in Supplementary file 2) out of Bambu PETG Translucent Clear. Both printers used a Smooth High Temp Plate. For explicit control, vascular structures were designed as two nested bodies: a distributed permeability infill and an encapsulation, both with a \SI{0.1}{\milli\meter} layer height.  The distributed permeability area consisted of 3 layers at 25\% rectilinear infill. The encapsulation consisted of two bottom layers and five top layers, totalling \SI{1}{\milli \meter} for the whole structure. The default (100\%) bridging layer density was used. The encapsulation was leakproofed by ironing each layer at 8\% ironing flow and  \SI{0.1}{\milli \meter} ironing line spacing.  Bambu Liquid Glue was used for promoting bed adhesion. Logistic highways (veins) were designed with a rectangular cross-section of \SI{1}{\milli \meter} width and \SI{1.4}{\milli \meter} height, positioned above the distributed permeability area to ensure consistent liquid access. Occasional minor encapsulation defects were corrected using a UV-curable glue. A pause command was inserted after the second layer for manual electrode placement. 

\textit{Receptor electrode placement procedure.} The electrodes consisted of a pair of \SI{30}{\micro \meter} gold wires (Surepure Chemetals LLC, USA) and placed using an in-house built wire placement device constructed by repurposing a CNC Kitchen hot melt insert installation adapter (TS100/ TS101 version) with a male M5 thread, combined with Pine64 soldering iron Pinecil V2 (Supplementary Figure S4). The base of the electrode placement device was an M5 brass hot melt insert. A custom sterling silver blade with a tip width of \SI{1}{\milli \meter}  was soldered to the insert. Two glass capillaries (BLAUBRAND intraMark BRAND 708733 \SI{50}{\micro \litre}) were melted together, drawn into a double cone, and trimmed to length.  The capillaries were bonded to a silver blade using CorroProtect Exhaust Paste.  The device was operated by feeding a gold wire pair through glass capillaries and guiding it by matching two indentations at the blade, ensuring approximately \SI{130}{\micro \meter} spacing. Careful manual operation of the heated \SI{195}{\celsius} blade with minimal pressing embedded the wires into the substrate polymer. This also resulted in the wires being over-molded by the molten polymer, raising them slightly above the surrounding substrate. The abrasive action of the indents kept the electrode tops exposed to the circulatory system.  For the thorax section, the electrodes were terminated outside the print area for external connection. For the wing, the wires were accessed via a printed-in passthroughs and fixed with conductive paint (MGChemicals, 843WB).

Vascularization was visualised with isopropanol (Honeywell, 33539), and 1:1 vol\% mixture of deionised water and isopropanol with a blue tracer (Pilc Pig Blue, Smooth-On).

\textit{UV Photopolymerisation and photolithography.} Pyrrole (Py, Sigma, 102394738) was distilled under reduced pressure and stored under argon at \SI{-18}{\celsius}. Cellulose acetate proprionate (CAP, avg Mn$\sim$ 15 000 by GPC, Aldrich 340642) was used to enhance mechanical properties \cite{Shi2014InRegeneration}. Bis(4-methylphenyl)iodonium hexafluorophosphate (98\%, 726192 Aldrich) was used as a photoinitiator, similarly to literature \cite{Rabek1992PolymerizationPhotoinitiators}. 15:30:1 weight ratio of Py:initiator:CAP was used. A 365-nm light-emitting diode (CUN66A1B, Seoul Viosys, Republic of Korea) source was used to photopolymerise Py with \SI{60}{\second} of illumination time. Negative contact masks for directing UV-light were made with aluminium tubing, 3D-printed holders (designed in Solidworks 2020, sliced in PrusaSlicer, and printed on a Prusa i3 MK3S+ printer with PETG), and aluminium tape. The LED was driven at 1.6 W and placed at \SI{43}{\milli\meter} from the sample. Molecular structures in Figure 3e were drawn using Ketcher (version 3.4.0)

Masks seen in Figure \ref{fig:3}d were designed in Autodesk Fusion and printed on Bambu A1 with \SI{0.2}{\milli \meter} nozzle and Prusa Research Original Prusa XL Five-Head with \SI{0.25}{\milli \meter} nozzles. The negative masks were printed with Prusa Research Prusament PETG Jet Black at \SI{0.2}{\milli\meter} single layer on a transparent substrate, with two \SI{0.1}{-\milli\meter} layers of Bambu PETG Translucent Clear. Both substrate layers were ironed. Minor corrections were made to the positive mask with a carving knife and black ink. 

\textit{Spectrometry.} Optical density of PPy-PETG was monitored \textit{in situ} during photopolymerization in transmission mode using Thorlabs CCS200/M spectrometer (specter collection time interval \SI{150}{\milli \second}, ThorSpectra software). In addition to the UV LED used to initiate synthesis, a separately controlled white LED was superimposed on the incident radiation to broaden the spectral range of optical density monitoring (without affecting synthesis). The integrated \SI{580}{\nano \meter} peak was used as a marker of the progression of synthesis. Data processing, including smoothing and normalization, was performed using National Instruments' LabVIEW 2020. 

\textit{Electrical characterization}. Biologic BP-300 electrochemical impedance analyzer was used to monitor impedance in PEIS mode, constant \SI{115}{\hertz}, and \SI{10}{\milli\volt} amplitude (Figure \ref{fig:3}g). For Figure 3f and 3g, supply current (Rigol DP832), of the UV LED was varied (\SI{0.43}{\ampere}, \SI{0.21}{\ampere}, \SI{0.1}{\ampere}, \SI{0.05}{\ampere}, \SI{0.25}{\ampere}, \SI{0.01}{\ampere}). Each \SI{10}{\second} rectangular current pulse was followed by a \SI{890}{\second} rest period. In parallel, Biologic BP-300 was used to monitor impedance response (PEIS mode, 115 Hz. Drift was corrected with a linear baseline. For Supplementary Figure S3, constant \SI{105}{\hertz} (Py-PETG) and constant \SI{120}{\hertz} (PPy-PETG) was used with \SI{10}{\milli\volt} amplitude, Savitzky-Golay filter (window size 21 and polynomial order 3) was applied with Python for data processing. Oscilloscope (Rigol DS1054) was used to capture transients in Supplementary Figure S2. 

\textit{Embedded receptor}. An embedded impedance measurement system was designed based on the ATtiny43U microcontroller clocked at \SI{1}{\mega \hertz}, programmed with an Atmel‑ICE in‑circuit debugger and MPLAB X IDE, and powered by an onboard 10-mAh single-cell lithium-polymer battery (\SI{4.2}{\volt} nominal). A voltage divider, formed by a linear variable resistor set to \SI{460}{\kilo \ohm} and the PPy, was applied intermittent alternating-polarity constant-voltage pulse pairs (two times \SI{200}{\milli \second}) using the processor's digital input-output pins. The pulse pairs were applied at \SI{1.4}{\second} intervals, and the outputs were switched to high-impedance state in between pulses to minimize receptor charging effects. The average of eight consecutive ratiometric voltage readings from the 10-bit ADC channel was used to approximate impedance.  The impedance readout was fed into a 3-s circular buffer. The buffer data was converted to impedance rate. 

The magnitude and sign of the impedance rate were indicated using LED‑based signaling. 'Signal detected' mode was enacted by four successive 4-Hz LED blinks, corresponding to the impedance rate exceeding a \SI{2}{\kilo \ohm} threshold corresponding to approx 1\% relative impedance drop in our sample, whereas the color of the active LED - red or yellow - indicated its sign (decrease or increase, respectively). A single \SI{0.5}{\second} simultaneous blink of both LEDs indicated the impedance rate was below the threshold. Count-based glitch filtering was implemented using a five-count circular buffer: five consecutive impedance-decrease readings triggered a \SI{65}{\second} 'reaction' mode that activated the wing-flapping mechanism (thorax receptor only), accompanied by the red LED blinking rapidly ten times.

\textit{Flapping wing mechanism.} Nitinol wire (nitinolxy \SI{0.5}{\milli \meter} \SI{45}{\celsius} transition temperature) was woven in 14 passes,  \SI{13}{\milli \meter} each, between the thorax and the wing sections. The binary logic-level control signal from the embedded receptor was applied to the gate of an n-type MOSFET (PHB66NQ03LT) that connected the Nitinol wire to an external power source (Rigol DP832), causing Nitinol to recover the programmed straight shape using electrothermal activation. One flap consisted of \SI{5}{\second} activation at \SI{6.6}{\watt}, followed by \SI{60}{\second} for natural convective cooling. A recoil spring and rubber bands aided the return to the starting position.

\textit{SEM imaging.} Hitachi TM3000 scanning electron microscope with back-scatter electron detector and 15-kV acceleration voltage was used for Figure \ref{fig:1}a inset and Figure \ref{fig:2}c, d. For Figure \ref{fig:1}a inset, the sample was cryofractured in liquid nitrogen to obtain a smooth cross-section, air-dried overnight, and then sputter-coated  (Leica EM ACE600 Sputter Coater, Wetzlar) with \SI{5}{\nano \meter} of gold. For Figure \ref{fig:2}c and d, prior to imaging, samples were washed using an isopropanol:water 1:1 by volume mixture, air‑dried, and subsequently cryofractured in liquid nitrogen. The fractured samples were then dried under a vacuum until the pressure reached 2-3 mPa and sputter‑coated with \SI{7}{\nano\meter} of gold (999).

\textit{Optical Imaging.} Sony FX30 and Sony A6300 were used for optical imaging and recording at 100 and 25 FPS, respectively. Adobe Photoshop was used to remove the background in Figure \ref{fig:4}a, d. Python code was used to extract receptor LED blinking from cropped video frames by first high-pass filtering the transient values of red or yellow hues to remove drift, and then thresholding to extract switching moments.

\textit{Biological Models.} The blue underwing (\textit{Catocala fraxini}) was chosen for its size and relative local abundance. Dead specimens were obtained from the preserved catches of the national biodiversity monitoring program, collected in summer 2023 from northeastern Estonia using a light trap. For Figure \ref{fig:1}a left, the specimen was dried in position and pinned. To visualise the wing veins and membrane (Figure \ref{fig:1}a right), the specimen was infused with silicone oil (5 cSt, Carl Roth) due to its high permeability through the exoskeleton and low evaporation rate. The scales were carefully removed using a paintbrush.

\textit{Large Language Models.} Microsoft Copilot (GPT-5.2; Microsoft Corporation; accessed December 2025–January 2026) and Anthropic Claude (Sonnet 4.6 and Opus 4.7; accessed April - May 2026) were employed as an assistive tool for embedded microprocessor programming and Python-based script development, including LED intensity analysis and data visualization. All AI-assisted outputs were critically reviewed, validated, and revised by the authors to ensure scientific accuracy and reproducibility.

\section*{Acknowledgements}
The authors thank Prof. Toomas Tammaru and Kadri Ude (Institute of Ecology and Earth Sciences, University of Tartu) for providing \textit{C. fraxini} specimens and scientific discussions. This research was supported by the Estonian Research Council grants PRG1498 and TEM-TA56.

\section*{Conflict of Interest}

The authors declare no conﬂict of interest.

\section*{Author Contributions Statement}

Conceptualization: K.A.P., E.S., I.M.; Data Curation: K.A.P., L.Z.; Funding Acquisition: T.T, A.A., I.M.; Investigation: K.A.P., L.Z., H.P., A.M.; Methodology: K.A.P, L.Z., H.P., U.J., E.S., I.M.; Project Administration: I.M.; Supervision: T.T., E.S., I.M.; Validation: K.A.P., L.Z., I.M.; Visualization: K.A.P., L.Z., H.P, E.S., I.M.; Writing - Original Draft: K.A.P, L.Z., I.M.; Writing - Review and Editing: K.A.P., L.Z., H.P., A.M., U.J., T.T., A.A., E.S., I.M.

\section*{Data Availability Statement}

All data needed to support the conclusions of this manuscript are included in the main text or Supplementary Information. Additional data could be made available from the corresponding authors K.A.P. and I.M. upon reasonable request.

\nocite{Gardner1970MATHEMATICALGAMES}

\bibliographystyle{unsrt}
\bibliography{references}

\end{document}